# DeepSF: deep convolutional neural network for mapping protein sequences to folds

Jie Hou[1], Badri Adhikari[1] and Jianlin Cheng[1,2,*]

[1]Department of Electrical Engineering and Computer Science, University of Missouri, Columbia, Missouri, 65211, USA.

[2]Informatics Institute, University of Missouri, Columbia, Missouri, 65211, USA.

*To whom correspondence should be addressed.

## Abstract

## Motivation

Protein fold recognition is an important problem in structural bioinformatics. Almost all traditional fold recognition methods use sequence (homology) comparison to indirectly predict the fold of a target protein based on the fold of a template protein with known structure, which cannot explain the relationship between sequence and fold. Only a few methods had been developed to classify protein sequences into a small number of folds due to methodological limitations, which are not generally useful in practice.

## Results

We develop a deep 1D-convolution neural network (DeepSF) to directly classify any protein sequence into one of 1195 known folds, which is useful for both fold recognition and the study of sequence-structure relationship. Different from traditional sequence alignment (comparison) based methods, our method automatically extracts fold-related features from a protein sequence of any length and map it to the fold space. We train and test our method on the datasets curated from SCOP1.75, yielding a classification accuracy of 80.4%. On the independent testing dataset curated from SCOP2.06, the classification accuracy is 77.0%. We compare our method with a top profile-profile alignment method - HHSearch on hard template-based and template-free modeling targets of CASP9-12 in terms of fold recognition accuracy. The accuracy of our method is 14.5%-29.1% higher than HHSearch on template-free modeling targets and 4.5%-16.7% higher on hard template-based modeling targets for top 1, 5, and 10 predicted folds. The hidden features extracted from sequence by our method is robust against sequence mutation, insertion, deletion and truncation, and can be used for other protein pattern recognition problems such as protein clustering, comparison and ranking.

## Availability

The DeepSF server is publicly available at: http://iris.rnet.missouri.edu/DeepSF/.

**Contact:** chengji@missouri.edu

## 1 Introduction

Protein folding reveals the evolutionary process between the protein amino acid sequence and its atomic tertiary structure [1]. Folds represent the main characteristics of protein structures, which describe the unique arrangement of secondary structure elements in the infinite conformation space [2, 3]. Several fold classification databases such as SCOP [2], CATH[4], FSSP[5] have been developed to summarize the structural relationship between proteins. With the substantial investment in protein structure determination in the past decades, the number of experimentally determined protein structures has substantially increased to more than 100,000 in the Protein Data Bank (PDB) [2, 6]. However, due to the conservation of protein structures, the number of unique folds has been rather stable. For example, the SCOP 1.75 curated in 2009 has 1195 unique folds, whereas SCOP 2.06 only has 26 more folds identified from the recent PDB [7]. Generally, determining the folds of a protein can be accomplished by comparing its structure with those of other proteins whose folds are known. However, because the structures of most (>99%) proteins are not known, the development of sequence-based computational fold detection method is necessary and essential to automatically assign proteins into fold. And identifying protein homologs sharing the same fold is a crucial step for computational protein structure predictions [8, 9] and protein function prediction [10].

Sequence-based methods for protein fold recognition can be summarized into two categories: (1) sequence alignment methods and (2) machine learning method. The sequence alignment methods [11, 12] align the sequence of a target protein against the sequences of template proteins whose folds are known to generate alignment scores. If the score between a target and a template is significantly higher than that of two random sequences, the fold of the template is considered to be the fold of the target. In order to improve the sensitivity of detecting remote homologous sequences that share the same fold, sequence alignment methods were extended to align the profiles of two proteins. Profile-sequence alignment method [13] and profile-profile alignment methods based hidden Markov model (HMM) [9] or Markov random fields (MRFs) [14] are more sensitive in recognize proteins that have the same fold, but little sequence similarity, than sequence-sequence alignment methods. Despite the success, the sequence alignment methods are essentially an indirect fold recognition approach that transfers the fold of the nearest sequence neighbors to a target protein, which cannot explain the sequence-structure relationship of the protein.

Machine learning methods have been developed to directly classify proteins into different fold categories [15-17]. However, because traditional machine learning methods cannot classify data into a large number of categories (e.g. thousands of folds), these methods can only classify proteins into a small number (e.g. dozens) of pre-selected fold categories, which cannot be generally applied to predict the fold of an arbitrary protein and therefore is not practically useful for protein structure predic-tion. To work around the problem, another kind of machine learning methods [8, 18, 19] converts a multi-fold classification problem into a binary classification problem to predict if a target protein and a template protein share the same fold based on their pairwise similarity features, which is still an indirect approach that cannot directly explain how a protein sequence is mapped to one of thousands of folds in the fold space.

In this work, we utilize the enormous learning power of deep learning to directly classify any protein into one of 1195 known folds. Deep learning techniques have achieved significant success in computer vision, speech recognition and natural language processing [20, 21]. The application of deep learning in bioinformatics has also gained the traction since 2012. Deep belief networks [22] were developed to predict protein residue-residue contacts. Recently a deep residual convolutional neural network was designed to further improve the accuracy of contact prediction [23]. Deep learning methods have also been applied to predict protein secondary structures [24, 25] and identify protein pairs that have the same fold [8, 14].

Here, we design a one-dimensional (1D) deep convolution neural network method (DeepSF) to classify proteins of variable-length into all 1195 known folds defined in SCOP 1.75 database. DeepSF can directly extract hidden features from any protein sequence of any length through convolution transformation, and then classify it into one of thousands of folds accurately. The method is the first method that can map all protein sequences in the sequence space directly into all the folds in the fold space without relying on pairwise sequence comparison (alignment). The hidden fold-related features generated from sequences can be used to measure the similarity between proteins, cluster proteins, and select template proteins for tertiary structure prediction.

We rigorously evaluated our method on three test datasets: new proteins in SCOP 2.06 database, template-based targets in the past CASP experiments, and template-free targets in the past CASP experiments. Our method (DeepSF) is more sensitive than a state-of-the-art profile-profile alignment

method – HHSearch in predicting the fold of a protein, and it is also much faster than HHSearch because it directly classifies a protein into folds without searching a template database. We also demonstrate that the hidden features extracted from protein sequences by DeepSF is robust against residue mutation, insertion, deletion and truncation.

## 2 Methods

### 2.1 Datasets

#### 2.1.1 Training and validation datasets

The main dataset that we used for training and validation was downloaded from the SCOP 1.75 genetic domain sequence subsets with less than 95% identity released in 2009. The protein sequences for each SCOP domain were cleaned according to the observed residues in the atomic structures [2]. The dataset contains 16,712 proteins covering 7 major structural classes with total 1,195 identified folds. The number of proteins in each fold is very uneven, with 5% (i.e. 61/1,195) folds each having >50 proteins, 26% (i.e. 314/1,195) folds each having 6 to 50 proteins, and 69% (820/1,195) each having <=5 proteins (supplemental Figure S1), making it challenging to train a classifier accurately predicting all the folds, especially small folds with few protein sequences. The proteins in all 1,195 folds have sequence length ranging from 9 to 1,419 (supplemental Figure S2 (a)), and most of them have length in the range of 9 to 600 (supplemental Figure S2 (b)).

The SCOP 1.75 dataset with less than or equal to 95% sequence identity was split into training and validation datasets with ratio 8/2 for each fold. Specifically, 80% of the proteins in each fold are used for training, and the remaining 20% of proteins are used for validation. The training proteins from all folds are combined to generate the final training dataset and likewise for the final validation dataset. The validation dataset was further filtered to at most 70%, 40%, 25% pairwise similarity with the training dataset for rigorous model selection and validation.

**2.1.2 Independent SCOP 2.06 test dataset**

In order to independently test the performance of our method, we collected the protein sequences in the latest SCOP 2.06 [7], but not in SCOP 1.75. The sequences with similarity greater than 40% with SCOP 1.75 dataset were further removed. And the remaining proteins were filtered to less than or equal to 40% pairwise similarity by CD-Hit suite [26]. Finally, this independent SCOP test dataset contains 4,188 domains, covering 550 folds.

**2.1.3 Independent CASP test dataset**

Besides classifying the proteins with known folds in the SCOP, we tested our methods on a protein dataset consisting of template-free and template-based targets used in the 9$^{th}$, 10$^{th}$, 11$^{th}$, and 12$^{th}$ Critical Assessments of Structure Prediction (CASP) experiments from 2010 to 2016 [27]. These are new proteins available after SCOP 1.75 was created in 2009. The complete CASP dataset contains 431 domains. The sequences in the CASP dataset with sequence identity >10% against the SCOP training dataset are removed. To assign the folds to these CASP targets, we compare each CASP target against all domains in SCOP 1.75 using the structural similarity metric - TM-score [28]. If a CASP target has TM-score above 0.5 with a SCOP domain, suggesting they have the same fold, the fold of the SCOP domain is transferred to the CASP target [29]. If the CASP target does not have the same fold with any SCOP domain, it is removed from the dataset. After preprocessing, the dataset has 186 protein targets with fold assignment, which include 96 template-free (FM) or seemly template-free (FM/TBM) targets and 90 template-based (TBM) targets.

**2.2 Input feature generation and label assignment**

We generated four kinds of input features representing the (1) sequence, (2) profile, (3) predicted secondary structure, and (4) predicted solvent accessibility of each protein. Each residue in a sequence is represented as a 20-dimension zero-one vector in which only the value at the residue index is marked as 1 and all others are marked as 0. The position-specific scoring matrix (PSSM) for each sequence is calculated by using PSI-BLAST to search the sequence against the 'nr90' database. The 20 numbers in the PSSM corresponding to each position in the protein sequence is used as features to represent the profile of amino acids at the position. We predicted 3-class secondary structure (Helix, Strand, Loop) and two-class solvent accessibility (Exposed, Buried) for each protein sequence using SCRATCH [30]. The secondary structure of each position is represented by 3 binary numbers with one of them as 1, indicating which secondary structure it is. Similarly, the solvent accessibility at each position is denoted by two binary numbers. In total, each position of a protein sequence is represented by a vector of 45 numbers. The whole protein is encoded by L × 45 numbers. Each sequence is assigned to a pre-defined fold index in the range of 0 ~ 1,194 denoting its fold according to SCOP 1.75 definition, which is the class label of the protein.

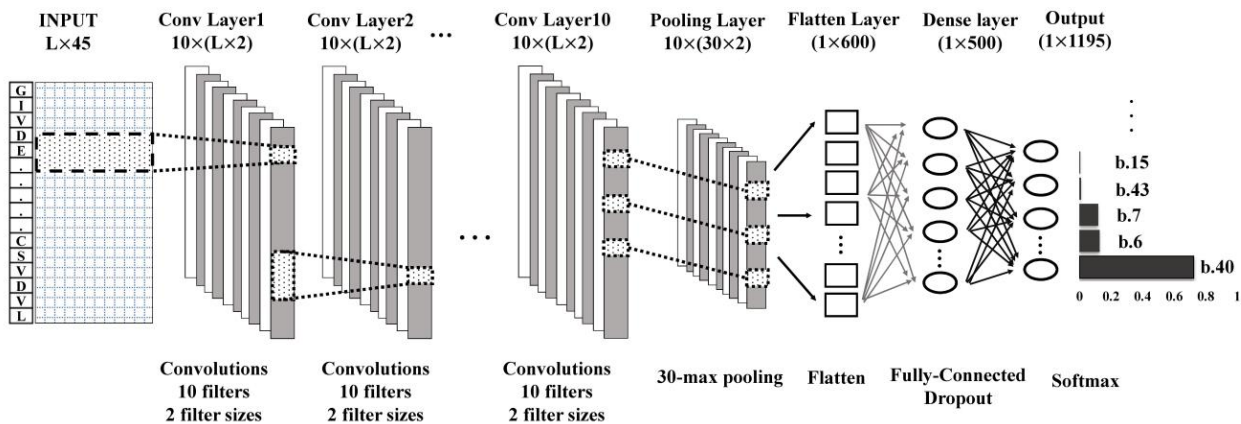

**Figure 1.** The architecture of 1D deep convolutional neural network for fold classification. The network accepts the features of proteins of variable sequence length (L) as input, which are transformed into hidden features by 10 hidden layers of convolutions. Each convolution layer applies 10 filters to the windows of previous layers to generate L hidden features. Two window sizes (6 and 10) are used. The 30 maximum values of hidden values of each filter of the 10$^{th}$ convolution layer are selected by max pooling, which are joined together into one vector by flattening. The hidden features in this vector are fully connected to a hidden layer of 500 nodes, which are fully connected to 1195 output nodes to predict the probability of each of 1195 folds. The output node uses softmax function as activation function, whereas all the nodes in the other layers use rectified linear function max(x, f(x)) as activation function. The features in the convolution layers are normalized by batches.

## 2.3 Deep convolutional neural network for fold classification

The architecture of the deep convolutional neural network for mapping protein sequences to folds (DeepSF) is shown in **Figure 1**. It contains 15 layers including input layer, 10 convolutional layers, one K-max pooling layer, one flattening layer, one fully-connected hidden layer and an output layer. The softmax function is applied to the nodes in the output layer to predict the probability of 1,195 folds. The input layer has $L \times 45$ input numbers representing the positional information of a protein sequence of variable length L. Each of 10 filters in the first convolution layer is applied to the windows in the input layer to generate $L \times 1$ hidden features (feature map) through the convolution operation, batch-normalization and non-linear transformation of its inputs with the rectified-linear unit (ReLU) activation function [20], resulting $10 \times L$ hidden features. Different window sizes (i.e. filter size) in the 1D convolution layer are tested and finally two window sizes (6 and 10) are chosen, which are close to the average length of beta-sheet and alpha-helix in a protein. The hidden features generated by 10 filters with two window sizes (i.e. $10 \times L \times 2$) in the first convolution layer are as input to be transformed by the second convolution layer in the same way. The depth of convolution layers is set to 10. Inspired by the work [31], the K-max pooling layer is added to transform the hidden features of variable length in the last convolution layer to the fixed number of features, where K is set to 30. That is the 30 highest values (30 most active features) of each $L \times 1$ feature map generated by a filter with a window size are extracted and combined. The extracted features learned from both window sizes (i.e., 6, 10) are merged into one single vector consisting of $10 \times 30 \times 2$ numbers, which is fed into a fully-connected hidden layer consisting of with 500 nodes. These nodes are fully connected to 1,195 nodes in the output layer to predict the probability of 1,195 folds. The node in the output layer uses the softmax activation function. To prevent the over-fitting, the dropout [32] technique is applied in the hidden layer (i.e. the $14^{th}$ layer in **Figure 1**).

## 2.4 Model training and validation

We trained the one-dimensional deep convolutional neural network (DeepSF) on variable-length sequences in 1,195 folds. Considering the proteins in the training dataset have very different length of up to 1,419 residues, we split the proteins into multiple mini-batches (bins) based on fixed-length interval (bin size). The proteins in the same bin have similar length in a specific range. The zero-padding is applied to the sequences whose length is smaller than the maximum length in the bin. All the mini-batches are trained for 100 epochs, and the proteins in each bin are used to train for a small number of epochs (i.e. 3 epochs for bin with size of 15) in order to avoid over-training on the proteins in a specific bin. We evaluated the performance of different bin sizes (see the Result section) to choose a good bin size. The DeepSF with different parameters is trained on the training dataset with less than or equal to 95% pairwise similarity, and is then evaluated on the validation sets with different sequence similarity levels (95%, 70%, 40%, 25%) with the training dataset. The model with the best average accuracy on the validation datasets is selected as final model for further testing and evaluation. A video demonstrating how DeepSF learns to classify a protein into a correct fold during training is available http://iris.rnet.missouri.edu/DeepSF/.

**2.5 Model evaluation and benchmarking**

We tested our method on the two independent test datasets: SCOP 2.06 (see Section 2.1.2) and CASP dataset (see Section 2.1.3). Since the number of proteins in different folds are extremely unbalanced, we split the 1,195 folds into three groups based on the number of proteins within each fold (i.e., small, medium, large). A fold is defined as 'small' if the number of proteins in the fold is less than 5, 'medium' if the number of proteins is in the range between 6 and 50, and 'large' if the number of proteins is larger than 50. We evaluated DeepSF on the proteins of all folds and those of each category in the test dataset separately. We compared DeepSF with the baseline majority-assignment method, which assigns the most frequent folds to the test proteins. Moreover, we compared DeepSF with a state-of-the-art profile-profile alignment method – HHSearch and PSI-BLAST on the CASP dataset based on top1, top5, top10 predictions, respectively.

**2.6 Hidden fold-related feature extraction and template ranking**

The outputs of the 14$^{th}$ layer of DeepSF (the hidden layer in fully connected layers) used to predict the folds can be considered as the hidden, fold-related features of an input protein, referred to as SF-Feature. The hidden features bridges between the protein sequence space and the protein fold space as the embedded word features connect a natural language sentence to its semantic meaning in natural language processing. Therefore, the hidden features extracted for proteins by DeepSF can be used to assess the similarity between proteins and can be used to rank template proteins for a target protein.

In our experiment, we evaluated the following four different distance (or similarity) metrics to measure the similarity between the fold-related features:

(1) Euclidean distance:

$$\text{Euclid-D:} \quad (Q, T) \mapsto \sqrt{\sum_{i=1}^{N}(Q_i - T_i)^2} \qquad (1)$$

(2) Manhattan distance:

$$\text{Manh-D:} \quad (Q, T) \mapsto \sum_{i=1}^{N}|Q_i - T_i| \qquad (2)$$

(3) Pearson's Correlation score:

$$\text{Corr-D:} (Q, T) \mapsto \log(1 - \text{Corr}(Q, T)) \qquad (3)$$

(4) KL-Divergence:

$$\text{KL-D:} (Q, T) \mapsto \sum_{i=1}^{N}(Q_i \log \frac{Q_i}{T_i} + T_i \log \frac{T_i}{Q_i}) \qquad (4)$$

where $Q, T$ is the SF-feature for query protein and template protein.

We randomly sampled 5 folds from the training dataset and sampled at most 100 proteins from the 5 folds to test the four metrics above. We use hierarchical clustering to cluster the proteins into 5 clusters, where the distance between any two proteins is calculated from their fold-related feature vectors by the four metrics, respectively. This process is repeated 1,000 times and the accuracy of clustering based on the four distance metrics are calculated and compared (see Results Section 3.3).

To select the best template for a target protein, the fold-related features of the target protein is compared with those of the proteins in the fold that the target protein is predicted to belong to. The templates in the fold are ranked in terms of their distance with the target protein.

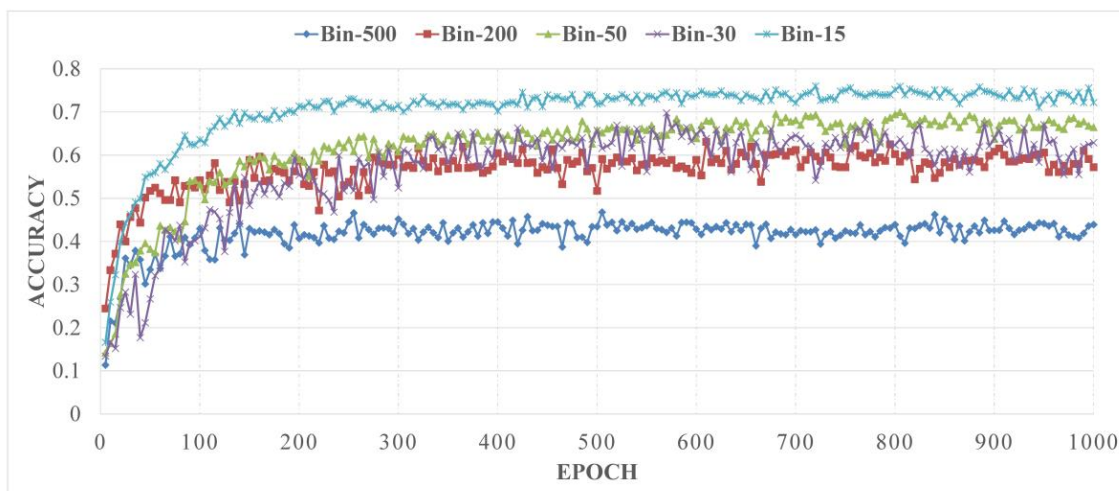

**Figure 2.** Classification accuracy of validation dataset against the number of training epochs for 5 different bin size.

## 3 Results

**3.1 Training and validation on SCOP 1.75 dataset**

We trained the deep convolutional neural network (DeepSF) on SCOP 1.75 dataset in the mini-batch mode, where the proteins in each mini-batch (bin) have similar length. We evaluated the effects of different bin sizes: 500, 200, 50, 30 and 15. The numbers of proteins within each batch (bin) are vis-ualized in supplemental Figure S3. The classification accuracy on the validation dataset with different bin sizes for each epoch of training is shown in **Figure 2**. Bin size of 15 has the fastest convergence and highest accuracy on both training (see Figure S4) and validation datasets, and therefore is chosen. The trained DeepSF classifiers with different parameter values were validated on four validation datasets that have at most 95%, 70%, 40% and 25% similarity with the training dataset. The accuracy of the best classifier is reported in **Table 1**. At the 95% similarity level, DeepSF achieves the accuracy of 80.4% (or 93.7%) for top 1 (or top 5) predictions. The classification accuracy decreases as the similarity level drops, reaching 66.9% (or 87.6%) at 25% similarity level for top 1 (or top 5) predictions. The average accuracy on all the four validation datasets is 75.3% (or 90.9%) for top 1 (or top 5) predictions.

**Table 1.** The prediction accuracy on four validation sets with different sequence similarity to training dataset for top 1, top 5, and top 10 predictions.

|        | ID < 95% | ID < 70% | ID < 40% | ID < 25% | Average |
|--------|----------|----------|----------|----------|---------|
| **Top 1**  | **80.4%** | **78.2%** | **75.8%** | **66.9%** | **75.3%** |
| **Top 5**  | 93.7%    | 92.4%    | 90.0%    | 87.6%    | 90.9%   |
| **Top 10** | 96.2%    | 95.4%    | 93.6%    | 92.1%    | 94.3%   |

**Table 2.** The accuracy of DeepSF on SCOP 2.06 dataset and its subsets.

| **DeepSF**         | Top1  | Top5  | Top10 |
|--------------------|-------|-------|-------|
| **SCOP2.06 dataset** | 76.9% | 92.2% | 95.7% |
| **"Large" folds**    | 82.9% | 95.6% | 98.1% |
| **"Medium" folds**   | 72.1% | 89.8% | 94.4% |
| **"Small" folds**    | 61.0% | 81.9% | 87.0% |

## 3.2 Performance on SCOP 2.06 dataset

We evaluated DeepSF on the independent SCOP 2.06 dataset, which contains 4,188 proteins belonging to 550 folds. 60 folds with 2,260 proteins are considered as "Large" fold, 249 folds with 1536 proteins as "Medium" fold and 241 folds with 392 proteins as "Small" fold. The classification accuracy of DeepSF on all the folds and each kind of fold is reported in **Table 2**. The accuracy on the entire dataset is 76.9% and 92.2% for top 1 prediction and top 5 predictions, respectively, much higher than the baseline majority-assignment method (top 1: 4.3%, top 5: 16.0 %), which demonstrates that DeepSF is significantly better than the simple statistics-based prediction. The model also achieves accuracy of 82.9%, 72.1% and 61.0% for top 1 prediction on "Large", "Medium", and "Small" folds, respectively. The higher accuracy on larger folds suggests that more training data in a fold leads to the better prediction accuracy.

## 3.2 Performance on CASP dataset

We evaluated our method on the CASP dataset, including 96 template-free proteins and 90 template-based proteins. We compared our method with the two widely used alignment methods (HHSearch and PSI-BLAST) as well as the majority-assignment method. Our method predicts the fold for each CASP target from its sequence directly. HHSearch and PSI-BLAST search each CASP target against the proteins in the training dataset to find the homologs to recognize its fold.

As shown in the **Table 3** and **Table 4**, DeepSF achieved better accuracy on both template-based targets and template-free targets than HHSearch, PSI-BLAST and majority-assignment in all situations. On the template-based targets that have little similarity with training proteins, the accuracy of DeepSF for top 1, 5, 10 predictions are 47.8%, 75.6%, 86.7% (see **Table 3**), which is 4.5%, 11.2%, 16.7% higher than HHSearch. And interestingly, the consensus ranking of HHSearch and DeepSF (Cons_HH_DeepSF) is better than both DeepSF and HHSearch, particularly for top 1 prediction, suggesting that the two methods are complementary on template-based targets. The majority assignment method sometimes has higher accuracy than HHSearch because 83 out of 90 template-based targets belong to 61 highly populated folds that are often chosen by the majority assignment. Because CASP targets has very low sequence similarity (<10%) with the training dataset, which is difficult for profile-sequence alignment methods to recognize, PSI-BLAST has the lowest prediction accuracy.

On the hardest template-free targets that presumably have no sequence similarity with the training dataset, the accuracy of DeepSF for top 1, 5 and 10 predictions are 26.0%, 57.3%, and 72.9% (see **Table 4**), 14.5%, 20.8% and 29.1% higher than HHSearch that performs better than PSI-BLAST. The consensus (Cons_HH_DeepSF) of DeepSF and HHSearch is only slightly better than DeepSF, which is different from its effect on template-based modeling targets. The majority assignment method performs better than HHSearch because 93 out of template-free targets happen to belong to 61 highly populated folds.

**Table 3.** The performance of methods on 90 template-based proteins in the CASP dataset

| Method | Top1 | Top5 | Top10 |
| --- | --- | --- | --- |
| DeepSF | **47.8%** | **75.6%** | **86.7%** |
| HHSearch | 43.3% | 64.4% | 70.0% |
| Cons_HH_DeepSF | 62.2% | 77.8% | 86.7% |
| Majority Assignment | 28.9% | 61.1% | 78.9% |
| PSI-Blast | 10.0% | 27.8% | 31.1% |

**Table 4.** The performance of methods on 96 template-free proteins in the CASP dataset

| Method | Top1 | Top5 | Top10 |
|---|---|---|---|
| DeepSF | **26.0%** | **57.3%** | **72.9%** |
| HHSearch | 11.5% | 36.5% | 43.8% |
| Cons_HH_DeepSF | 27.1% | 59.4% | 72.9% |
| Majority Assignment | 16.7% | 54.2% | 59.4% |
| PSI-Blast | 3.1% | 20.8% | 24.0% |

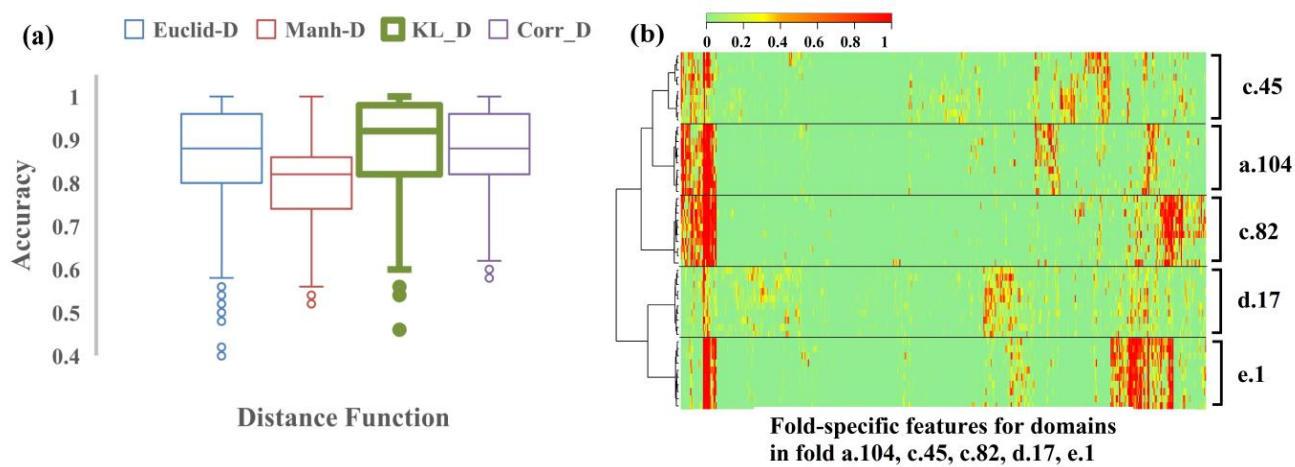

**Figure 3. (a)** The accuracy of 4 distance metrics in clustering proteins based on fold-related features. The clustering accuracy is average over 1000 clustering processes. **(b)** A hierarchical clustering of proteins from 5 folds in the SCOP 2.06 dataset using KL-D as metric. Each row in the heat map visualizes a vector of fold-related hidden features of a protein. The feature vectors of the proteins of the same fold are similar and clustered into the same group.

**3.3 Evaluation of four distance metrics for comparing fold-related hidden features**

We evaluated the four distance metrics by using hierarchical clustering to cluster proteins with known folds based on their hidden fold-related features (see Method Section 2.6). The boxplot in **Figure 3 (a)** shows the clustering accuracy of 4 different distance metrics. While Euclid-D, Manh-D and Corr-D achieve accuracy of 86.3%, 80.4%, and 88.0%, KL-D performs the best with accuracy of

89.3%. **Figure 3 (b)** shows an example that using KL-D as distance metric to cluster the fold-level features of proteins in five SCOP2.06 folds that are randomly sampled. The proteins are perfectly clustered into 5 groups with the same folds. The visualized heat map (**Figure 3 (b)**) shows that proteins in the same cluster (fold) has the similar hidden feature values. More detailed information including the name and SCOP id of the proteins is illustrated in supplemental Figure S5.

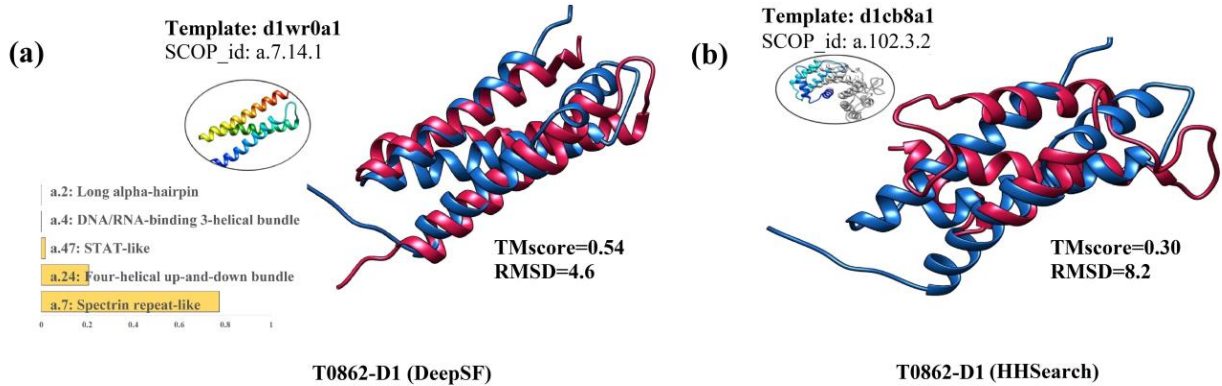

**Figure 4.** Tertiary structure prediction for CASP12 target T0862-D1 based on templates identified by DeepSF and HHSearch. **(a)** DeepSF predictions: a top template, five predicted folds and the supposition between the best model and the template structure; **(b)** HHSearch predictions: top template, and superposition of the best model and the template structure.

### 3.4 Fold-classification assisted protein structure prediction

Since applying a distance metric such as KL-D to the fold-related hidden features of two proteins can be used to measure their structural similarity, we explored the possibility of using it to rank template proteins for a target protein to assist tertiary structure prediction. Using the DeepSF model, we can generate fold-related features (SF-features) for any protein in a template protein database. In our experiment, we use DeepSF to generate SF-features for all the proteins in the training dataset as the template database. Given a target protein, we first extracted its SF-features and predicted the top 5 folds for it. We selected top 5 folds because top 5 predictions generally provided the high accuracy of fold prediction. Then we collected the template proteins that belong to the predicted top 5 folds and compare their SF-features with that of the target protein using KL-D metric. The templates are then ranked by KL-D scores from smallest to largest, and the top ranked 10 templates are selected to build the protein structures for the target proteins [33]. This method contrasts with the approach of HHSearch, where the target sequence is searched against the template database, and the top ranked 10 templates with smallest e-value are selected as candidate templates for protein structure prediction.

After the templates are detected by DeepSF or HHSearch, the sequence alignment between the target protein and each template are generated using HHalign [9]. Each alignment and its corresponding template structure are fed into Modeller [34] to build the tertiary structures. The predicted structural model with highest TMscore among all the models generated by top templates is selected for comparison. The quality of best predicted models from DeepSF and HHSearch is evaluated against the native structure in terms of TM-score and RMSD [28].

Here, we mainly evaluated template ranking and protein structure prediction on the 96 template-free CASP targets assuming that our method is more useful for detecting structural similarity for hard targets without sequence similarity with known templates. **Table 5** reports the average, min, max and standard deviation (std) of TMscore of the best models predicted for 96 template-free targets by DeepSF and HHSearch. DeepSF achieved a higher average TMscore (0.27) than that (0.25) of HHSearch. And the p-value of the difference using Wilcoxon paired test is 0.019. **Figure 4** shows an example on which DeepSF performed well. T0862-D1 is a template-free target in CASP 12, which contains multiple helices. DeepSF firstly classifies T0862-D1 into fold 'a.7' with probability 0.77 which is a 3-helix bundle. And among the top 10 ranked templates with smallest KL-D score in the fold 'a.7', the domain 'd1wr0a1' (SCOP id: a.7.14.1) was used to generate the best structural model with TMscore = 0.54 and RMSD = 4.6 Angstrom. In contrast, among the top 10 predicted structural models from HHSearch, the best model was constructed from a segment (residues 5-93) of a large template 'd1cb8a1' (SCOP id: a.102.3.2), which has TMscore of 0.30 and RMSD of 8.2.

## 3.5 Robustness of fold-related features against sequence mutation, insertion, deletion and truncation

In the evolutionary process of proteins, amino acid insertion, deletion or mutations mostly modifies protein sequences without changing the structural fold. Protein truncation that shortens the protein sequences at either N-terminal or C-terminal sometimes still retains the structural fold [35]. A good method of extracting fold-related features from sequences should capture the consistent patterns despite of the evolutionary changes. Therefore, we simulated these four residue changes to check if the fold-related features extract from protein sequences by DeepSF are robust against mutation, insertion, deletion and even truncation. To analyze the effects of mutation, insertion, and deletion, we selected some proteins that have 100 residues, and randomly selected the positions for insertion, dele-tion, or substitution with one or more residues randomly sampled from 20 standard amino acids. And at most 20 residues in total are deleted from or inserted into sequences. Each change was repeated 50 times, and the exactly same sequences were removed after sampling. For example, for domain d1lk3h2 we generated 44 sequences with at least one residue deleted, and 44 sequences with at least one residue insertion, and 18 sequences with at least one residue mutation. The SF-Features for these mutated sequences are generated and compared to the SF-Feature of the original wild-type sequence. We also randomly sampled 500 sequences with length in the range of 80 to 120 residues from the SCOP 1.75 dataset as control, and compare their SF-features with those of the original sequence. The distribution of KL-D divergences between the SF features of these sequences and the original sequence are shown in **Figure 5**. The divergence of the sequences with mutations, insertions, and deletions from the original sequence is much smaller than that of random sequences. The p-value of difference according to Wilcoxon rank sum test is < 2.2e-16. The same analysis is applied to the other two proteins: 'd1foka3' and 'd1ipaa2', and the same phenomena has been observed (see supplemental Figure S6). The results suggest that the feature extraction of DeepSF is robust against the perturbation of sequences.

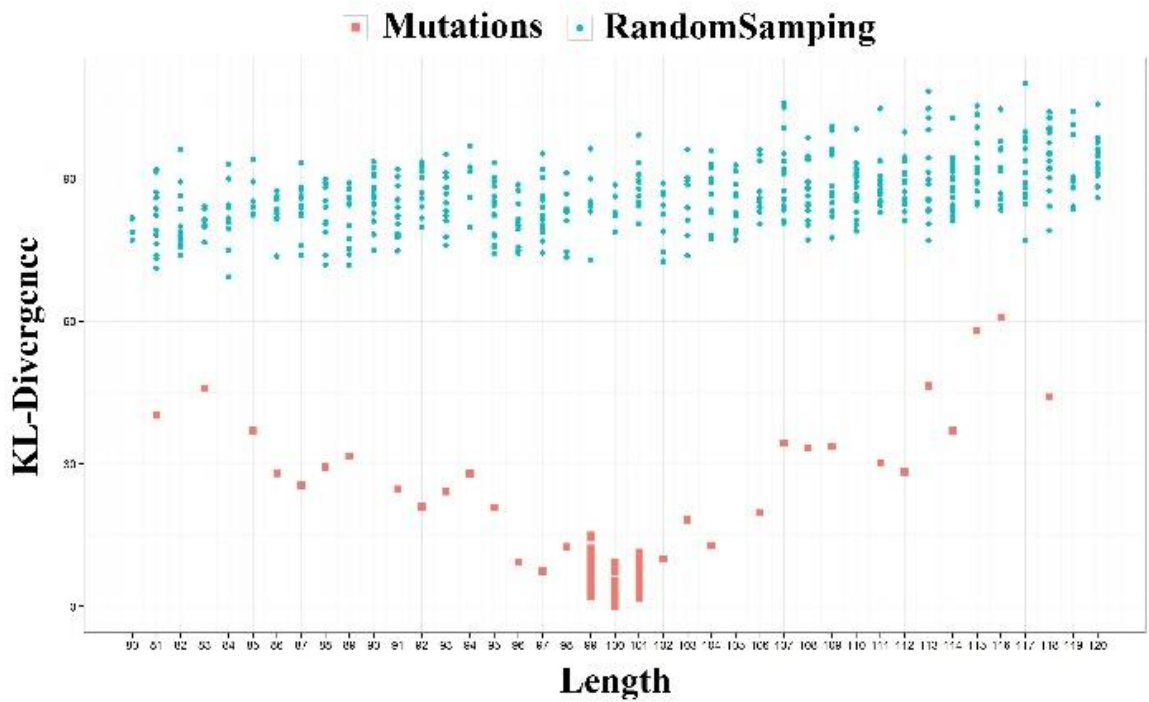

**Figure 5.** The KL-D divergences of fold-related features of 106 modified sequences of protein d1lk3h2 from the wild-type sequence (red dots) and those of 500 random sequences from the wild-type sequence (blue dots).

For the truncation analysis, we simulated residue truncations on C-terminus of 4,188 proteins in the SCOP 2.06 datasets by letting DeepSF read each protein's sequence from N-terminal to C-terminal to predict its fold. DeepSF needs to read 67.1% of the original sequences from N- to C-terminal on average in order to predict the same fold as using the entire sequences. This may suggest that the feature extraction is robust against the truncation of residues at C-terminal. A video demonstrating how DeepSF reads a protein sequence from N- to C-terminal to predict fold is available at http://iris.rnet.missouri.edu/DeepSF/.

# 4 Conclusion

We presented a deep convolution neural network to directly classify a protein sequence into one of all 1,195 folds defined in SCOP 1.75. To our knowledge, this is the first system that can directly classify proteins from the sequence space to the entire fold space rather accurately without using sequence comparison. Our method can automatically extract a set of fold-related hidden features from protein sequence of any length by deep convolution, which is different from previous machine learning methods relying on a window of fixed size or human expertise for feature extraction. The automatically extracted features are robust against sequence perturbation and can be used for various protein data analysis such as protein comparison, clustering, template ranking and structure prediction. And on the independent test datasets, our method is more accurate in recognizing folds of target proteins that have little or no sequence similarity with the proteins having known structures than widely used profile-profile alignment methods. Moreover, our method of directly assigning a protein sequence to a fold is not only complementary with traditional sequence-alignment methods based on pairwise comparison, but also provides a new way to study the protein sequence-structure relationship.

# 5 Acknowledgements

This work is partially supported by NIH R01 (R01GM093123) grant to JC

**Supplementary Figures**

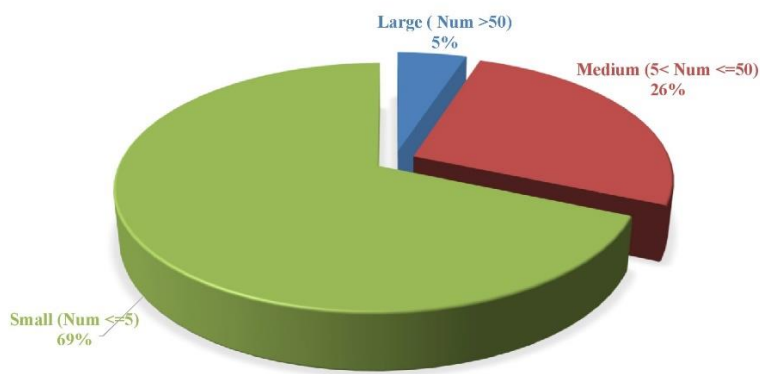

**Figure S1.** The proportion of three pre-defined groups for 1,195 folds in the training dataset. A fold is defined as 'Small' if the number of proteins in the fold is less than 5, 'Medium' if the number of proteins in the fold is in the range between 6 and 50, and 'Large' if the number of proteins is larger than 50.

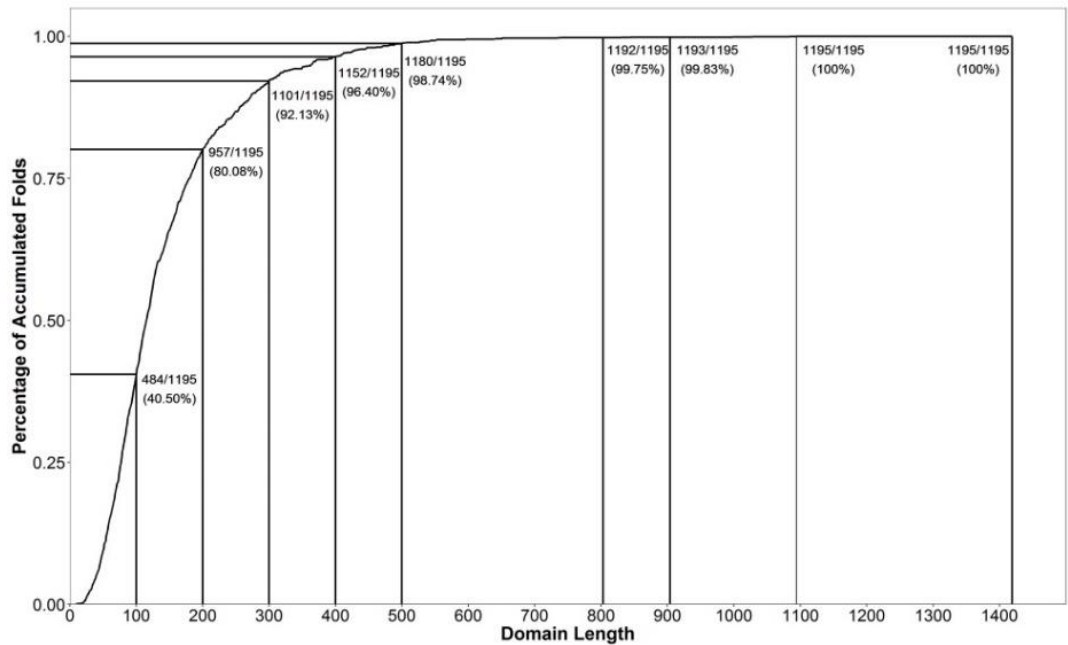

**Figure S2 (a).** The percentage of accumulated folds against length of proteins in the SCOP 1.75 dataset. In this plot, all the proteins with length less than 1,419 contains all 1,195 folds.

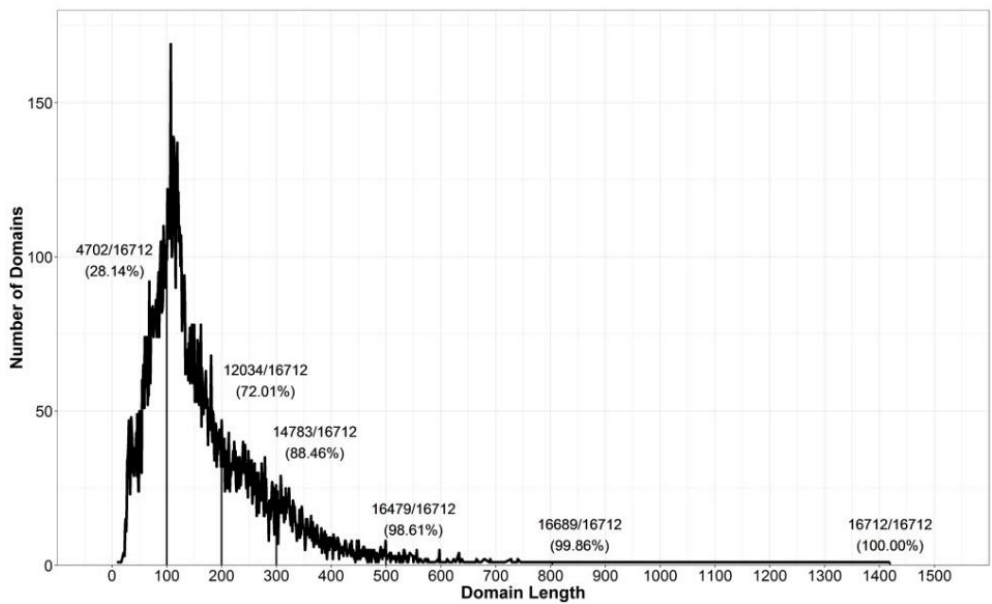

**Figure S2 (b).** The distribution of the number of domains versus length of proteins in the SCOP 1.75 dataset. The proteins in SCOP 1.75 dataset with sequence similarity at most 95% have sequence length ranging from 9 to 1,419.

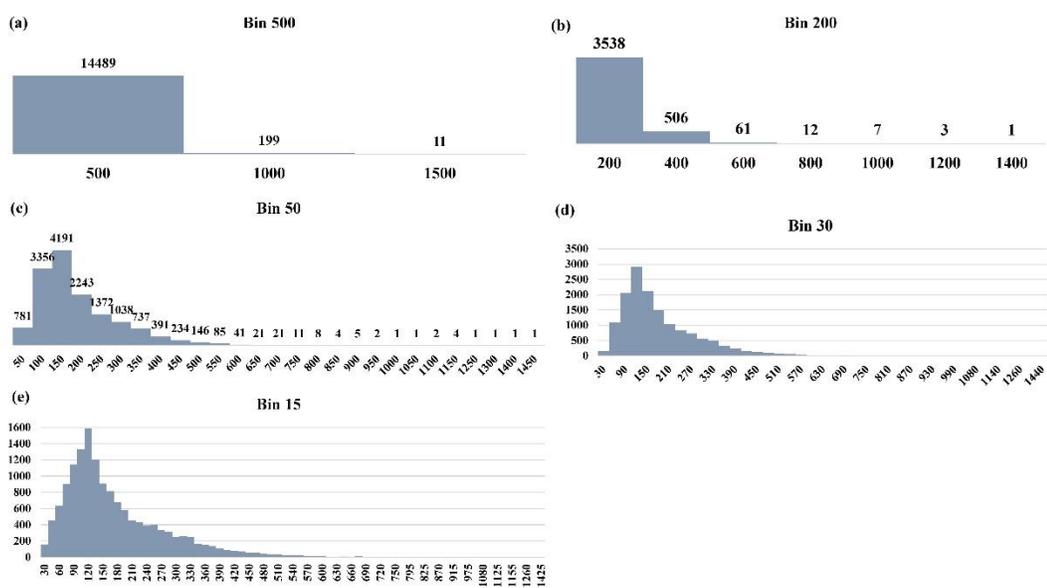

**Figure S3.** The distribution of the numbers of proteins within each batch with bin size as 500, 200, 50, 30, 15. The x axis denotes the length interval of mini-batches, and y axis denotes the number of proteins in the mini-batch.

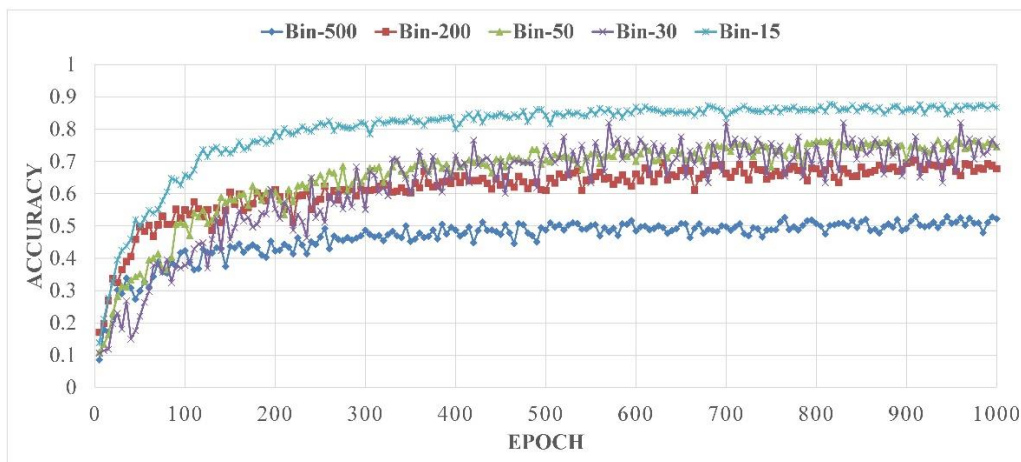

**Figure S4.** The Classification accuracy of training dataset against the number of training epochs for 5 different bin size.

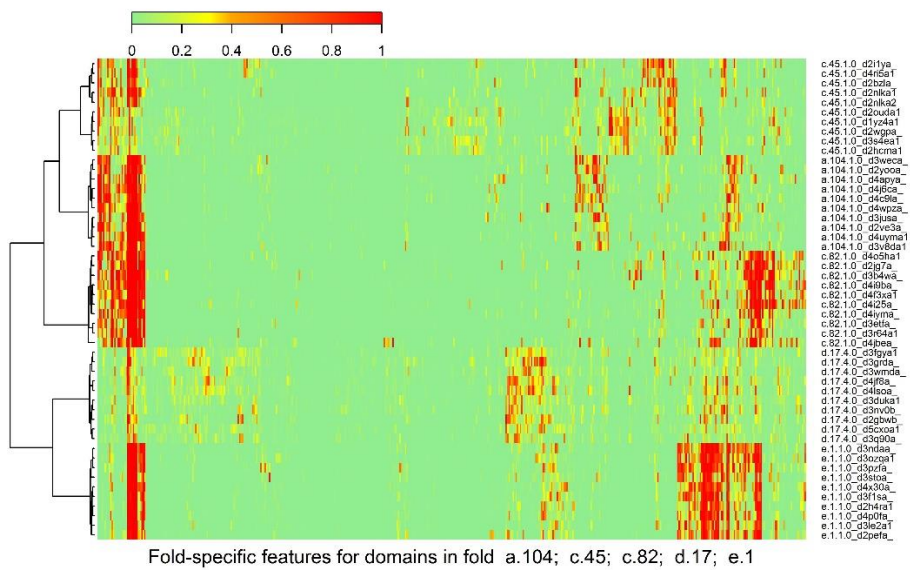

Fold-specific features for domains in fold a.104; c.45; c.82; d.17; e.1

**Figure S5.** The heatmap of SF-features of proteins from 5 folds. The features are clustered by Hierarchical clustering, and proteins in the same cluster (fold) has the similar hidden feature values.

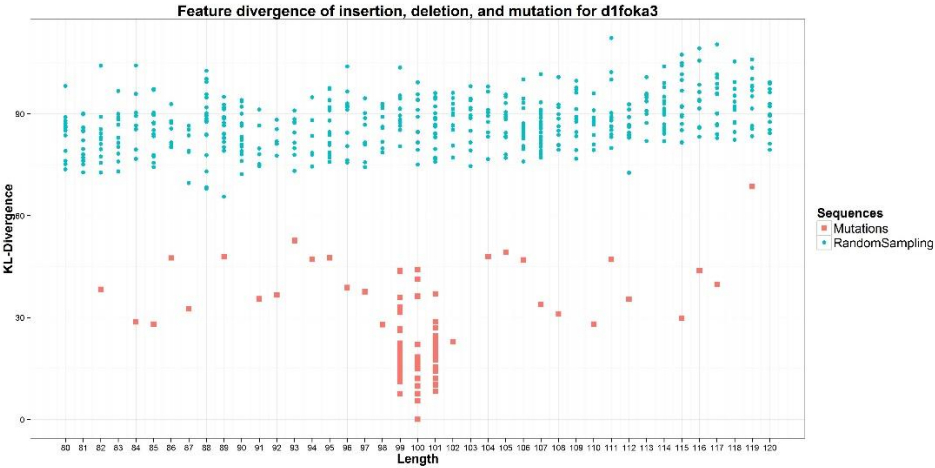

**Figure S6 (a).** The KL-D divergences of fold-related features of 102 modified sequences of protein d1foka3 from the wild-type sequence (red dots) and those of 500 random sequences from the wild-type sequence (blue dots). We generated 46 sequences with at least one residue deleted, and 40 sequences with at least one residue insertion, and 16 sequences with at least one residue mutation.

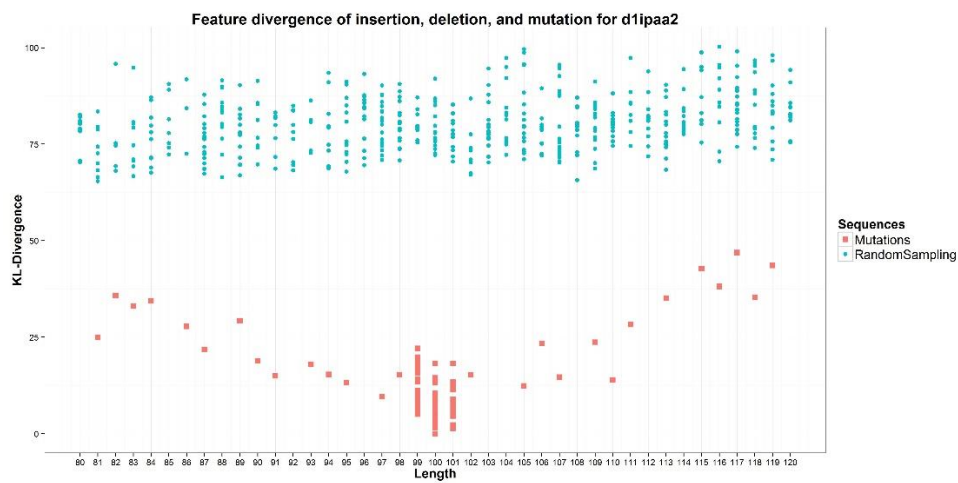

**Figure S6 (b).** The KL-D divergences of fold-related features of 106 modified sequences of protein d1ipaa2 from the wild-type sequence (red dots) and those of 500 random sequences from the wild-type sequence (blue dots). We generated 45 sequences with at least one residue deleted, and 41 sequences with at least one residue insertion, and 20 sequences with at least one residue mutation.